\pdfoutput=1

\documentclass[11pt]{article}

\usepackage[]{acl}

\usepackage{times}
\usepackage{latexsym}

\usepackage{graphicx}
\usepackage{subfigure}
\usepackage{booktabs}
\usepackage{multirow}
\usepackage{amsthm,amsmath,amssymb}
\usepackage{mathrsfs}
\usepackage{colortbl}
\usepackage{arydshln}
\usepackage{float}
\usepackage[normalem]{ulem}
\definecolor{mygray}{gray}{.95}

\usepackage[T1]{fontenc}

\usepackage[utf8]{inputenc}

\usepackage{microtype}

\title{Graph Pre-training for {AMR} Parsing and Generation}

\author{
 Xuefeng Bai$^{\spadesuit \heartsuit}$\hspace{0.5mm}, 
 Yulong Chen$^{\spadesuit \heartsuit}$\hspace{0.5mm}, 
 Yue Zhang$^{\heartsuit \diamondsuit}$\hspace{0.2mm}\hspace{1.5mm} \\
 $^\spadesuit$ Zhejiang University, China\\
 $^\heartsuit$ School of Engineering, Westlake University, China\\
 $^\diamondsuit$ Institute of Advanced Technology, Westlake Institute for Advanced Study, China
}

\begin{document}
\maketitle
\begin{abstract}
Abstract meaning representation (AMR) highlights the core semantic information of text in a graph structure.
Recently, pre-trained language models (PLMs) have advanced tasks of AMR parsing and AMR-to-text generation, respectively.
However, PLMs are typically pre-trained on textual data, thus are sub-optimal for modeling structural knowledge.
To this end, we investigate graph self-supervised training to improve the structure awareness of PLMs over AMR graphs.
In particular, we introduce two graph auto-encoding strategies for graph-to-graph pre-training and four tasks to integrate text and graph information during pre-training.
We further design a unified framework to bridge the gap between pre-training and fine-tuning tasks.
Experiments on both AMR parsing and AMR-to-text generation show the superiority of our model.
To our knowledge, we are the first to consider pre-training on semantic graphs.

\end{abstract}

\section{Introduction}
\begin{figure}
    \centering
    \includegraphics[width=0.45\textwidth]{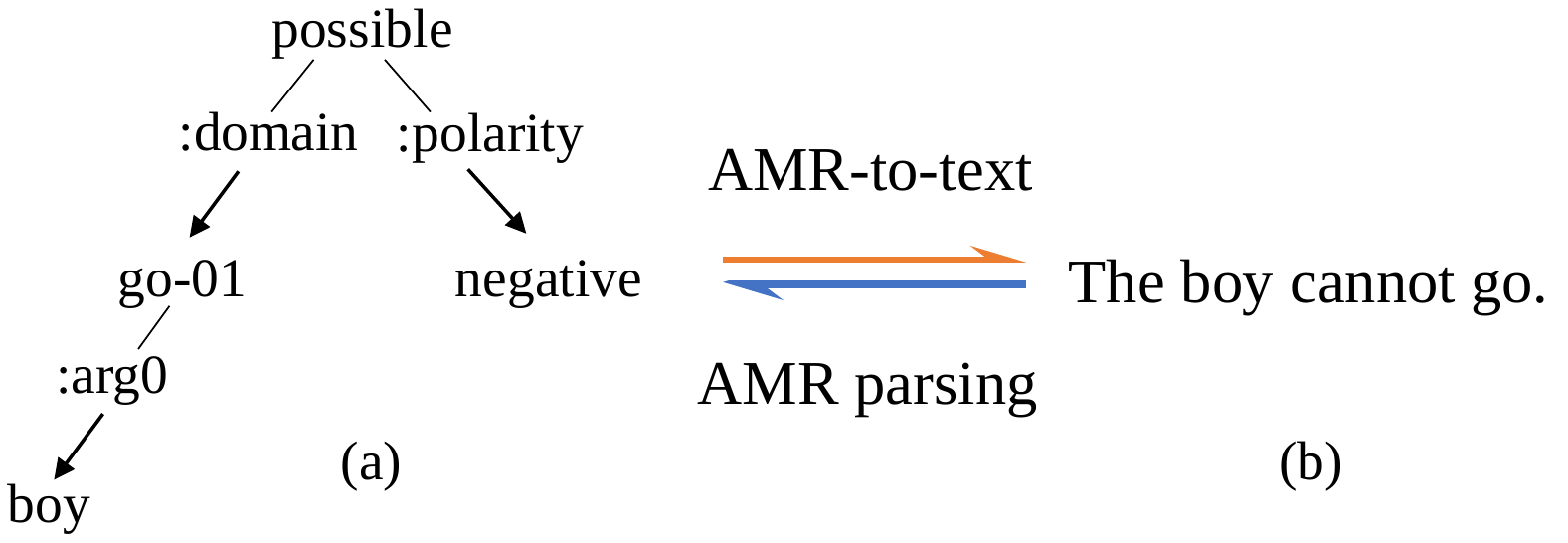}
    \caption{Illustration of AMR tasks: (a) an AMR graph; (b) a corresponding sentence.}
    \label{fig:intro-example}
\end{figure}

Abstract meaning representation (AMR; \citet{banarescu2013abstract}) is a semantic structure formalism. 
It represents the meaning of a text in a rooted directed graph, where nodes represent basic semantic units such as entities and predicates, and edges represent their semantic relations, respectively.
One example is shown in Figure~\ref{fig:intro-example}(a), with the corresponding sentence in Figure~\ref{fig:intro-example}(b). Serving as a structural representation, AMR has been shown useful for NLP tasks such as text summarization \cite{liu-etal-2015-toward,liao2018abstract,chen-etal-2021-dialogsum}, machine translation \cite{song2019semantic}, information extraction~\cite{huang-etal-2016-liberal,zhang-ji-2021-abstract} and dialogue systems \cite{bai-etal-2021-semantic}.

There are two fundamental NLP tasks concerning AMR, namely AMR parsing \cite{flanigan-etal-2014-discriminative,konstas2017neural,TitovL18,guo-lu-2018-better,zhang-etal-2019-amr,cai-lam-2020-amr,Bevilacqua_Blloshmi_Navigli_2021} and AMR-to-text generation \cite{konstas2017neural,song2018graph,zhu2019modeling,zhao-etal-2020-line,bai-etal-2020-online,ribeiro-etal-2021-investigating}.
As shown in Figure 1, the former transforms a textual input (\emph{e.g.}, a sentence) into a corresponding AMR structure, and the latter transforms an AMR input into a fluent and grammatical sentence that conveys the same meaning. A common challenge to both tasks is that AMR exists in the form of a graph structure, which is difficult for neural models to learn with limited human-curated data.

Recently, large-scale pre-trained sequence-to-sequence (seq2seq) language models \cite{lewis-etal-2020-bart,JMLR:v21:20-074} have been shown useful for both tasks above. 
The basic idea is to linearize AMR structures into a sequence form, so that both AMR parsing and AMR-to-text generation can be solved as standard seq2seq tasks, using a pre-trained language model fine-tuned on task-specific data. 
In this way, semantic knowledge learned in self-supervised text-to-text (\texttt{t2t}) pre-training can benefit both text-to-graph (\texttt{t2g}) and graph-to-text (\texttt{g2t}) transformation.

Intuitively, structural knowledge from AMR can be a useful complement to semantic knowledge from text. A natural question is whether similar self-supervision strategy can be useful for AMR graphs, so that graph-to-graph (\texttt{g2g}) denoise auto-encoder training can serve as effective addition to \texttt{t2t} pre-training, before a model is fine-tuned on \texttt{t2g} and \texttt{g2t} tasks. We investigate this problem in this paper.
In particular, there are three questions of interest. 
\textit{First, as mentioned before, is \texttt{g2g} pre-training complementary to \texttt{t2t} pre-training?}
\textit{Second, what is the most effective way to combine \texttt{t2t} and \texttt{g2g} training?} 
\textit{Third, is silver data useful for AMR self-supervision training, and what is the most effective way of making use of such data?} 

Taking BART~\cite{lewis-etal-2020-bart} as the seq-to-seq model, we introduce two strategies for \texttt{g2g} pre-training and propose four tasks to combine \texttt{t2t} and \texttt{g2g} training.
To reduce the gap among different pre-training tasks and between pre-training and fine-tuing, we unify all pre-training tasks and fine-tuning tasks in a general framework.
Experimental results on standard benchmarks show that: 1) graph pre-training achieves significant improvements over the state-of-the-art systems; 
2) silver data are useful for our pre-training framework;
3) our pre-training framework is a better way than fine-tuning to make use of silver data and;
4) our model is more robust than existing systems in unseen domains.
Our final models give the best reported results on both AMR parsing and AMR-to-text generation tasks, with a large margin of improvement over the previous best results.
To our knowledge, we are the first to consider graph-to-graph self-supervised training on semantic graphs.
We release code at~\url{https://github.com/muyeby/AMRBART}.

\section{Related Work}
\textbf{AMR Parsing.}
Early AMR parsing systems use statistical methods~\cite{flanigan-etal-2014-discriminative,flanigan-etal-2016-cmu,wang-etal-2015-boosting,wang-etal-2015-transition}. 
With the advance in deep learning, various neural models are developed for AMR parsing. Those models can be categorized into: 1) neural transition-based parsers~\cite{ballesteros-al-onaizan-2017-amr,liu-etal-2018-amr,fernandez-astudillo-etal-2020-transition,zhou-etal-2021-amr}; 2) sequence-to-graph parsers~\cite{zhang-etal-2019-amr,DBLP:journals/corr/abs-2010-12676,cai-lam-2020-amr} and; 3) sequence-to-sequence parsers~\cite{konstas2017neural,peng-etal-2017-addressing,peng-etal-2018-sequence,zhang-etal-2019-broad,xu-etal-2020-improving,Bevilacqua_Blloshmi_Navigli_2021}.
Recently, pre-training techniques have significantly boosted the performance of AMR parsing. 
For example, ~\citet{TitovL18},~\citet{zhang-etal-2019-amr,zhang-etal-2019-broad} and~\citet{cai-lam-2020-amr} use BERT~\cite{devlin-etal-2019-bert} for sentence encoding;
~\citet{Bevilacqua_Blloshmi_Navigli_2021} fine-tune BART for sequence-to-AMR generation.
~\citet{xu-etal-2020-improving} pre-train a model on relevant seq2seq learning tasks (e.g., machine translation~\cite{Bahdanau2015NeuralMT}, syntactic parsing~\cite{zhu-etal-2013-fast}) before fine-tuning on AMR parsing.
Similar to those methods, we consider using pre-trained models to improve the model capacity. 
However, previous studies focus on fine-tuning language models trained on text data for AMR parsing task, in contract, we focus on integrating structural information into the pre-training.
In addition, our method does not require information from auxiliary tasks.

\noindent\textbf{AMR-to-Text Generation.}
On a coarse-grained level, we categorize existing AMR-to-text generation approaches into two main classes:
Graph-to-sequence models that adopt a graph encoder to process an AMR graph and use a sequence decoder for generation~\cite{beck2018graph,damonte-cohen-2019-structural,zhu2019modeling}, and sequence-to-sequence models that linearize an AMR graph into a sequence and solve it as a seq2seq problem using randomly initialized~\cite{konstas2017neural} or pre-trained models~\cite{mager-etal-2020-gpt,ribeiro-etal-2021-investigating,Bevilacqua_Blloshmi_Navigli_2021}.
This work follows a seq2seq manner, but we use an encoder that integrates AMR and text information.
The closest to our work, ~\citet{Ribeiro2021StructuralAI} integrate AMR structures into pre-trained T5~\cite{JMLR:v21:20-074}  using adapters~\cite{Houlsby2019ParameterEfficientTL} for AMR-to-text generation.
However, they do not pre-train on AMR graphs, and their method cannot solve both AMR parsing and AMR-to-text generation tasks as they require the full AMR structure as the input.

\begin{figure*}[!t]
	\centering
	\includegraphics[width=0.9\textwidth]{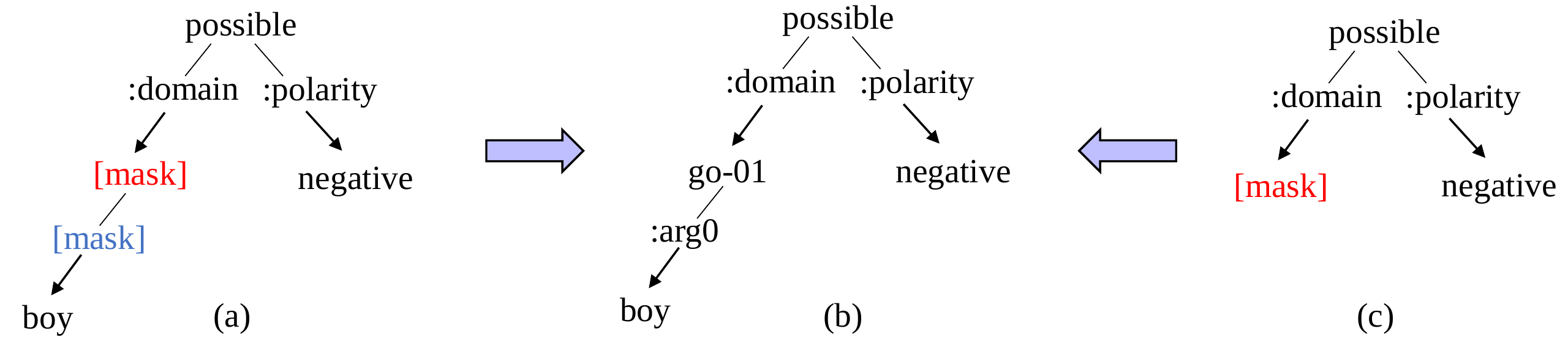}
	\caption{Illustration of two graph pre-training strategies: 1) node/edge level denoising (a$\rightarrow$ b); 2) sub-graph level denoising (c$\rightarrow$ b). Two transformations can be composed.}
	\label{fig:graph-masking}
\end{figure*}

\noindent\textbf{Graph Self-supervised Learning.}
~\citet{Kipf2016VariationalGA} introduce a variational graph auto-encoder to allow self-supervised learning on graph data.
~\citet{Hu2020StrategiesFP,Hu2020GPTGNNGP} propose local and global learning strategies to pre-train a graph neural network on large-scale protein ego-networks, academic graphs and recommendation data. 
~\citet{Lu2021LearningTP} enhance the graph learning strategies of ~\citet{Hu2020GPTGNNGP} with dual adaptations.
While existing work considers graph neural networks, we pre-train a seq2seq model on AMR graphs. 
In addition, we jointly pre-train on graphs and text for graph-text correlation modeling.
In contrast, existing work pre-trains models on graphs and in isolation with text pre-training.
To our knowledge, we are the first to consider AMR as a graph pre-training target.

\section{Method}
We take BART~\cite{lewis-etal-2020-bart} as the basic seq2seq model (Section~\ref{sec:bart}), and introduce graph pre-training strategies (Section~\ref{sec:graphrec}) and an unified pre-training framework (Section~\ref{sec:dualrec}) for both AMR parsing and AMR-to-text generation.
\subsection{BART}
\label{sec:bart}
BART~\cite{lewis-etal-2020-bart} is a pre-trained denoising auto-encoder, which is implemented as a seq2seq model based on standard Transformer~\cite{vaswani2017attention} architecture.
Typically, BART is trained to reconstruct original text based on a corrupted text generated by $5$ noising functions: 1) Token Masking. Tokens are randomly replaced by \texttt{[mask]} elements;
2) Token Deletion. Tokens are randomly deleted from the input;
3) Text Infilling. Text spans are randomly replaced by a single \texttt{[mask]} token;
4) Sentence Permutation. Text is divided into segments and then shuffled;
5) Document Rotation. A document is rotated to start with a random token.
In the fine-tuning, BART takes a complete text as input and maps it into a task-specific output sequence.

We linearize an AMR graph into a sequence, so that both AMR parsing and AMR-to-text generation can be performed using a seq2seq model.
In addition, it allows pre-training on AMR structures using BART.
Following \citet{konstas2017neural}, we adopt the depth-first search (DFS) algorithm, which is closely related to the linearized natural language syntactic trees~\cite{Bevilacqua_Blloshmi_Navigli_2021}.
For instance, the AMR graph in Figure~\ref{fig:intro-example} is linearized into:
\texttt{( <Z0> possible :domain ( <Z1> go :arg0 ( <Z2> boy ) ) :polarity ( <Z3> negative ) )} ,

\noindent where \texttt{<Z0>}, \texttt{<Z1>} and \texttt{<Z2>} are special tokens to handle co-referring nodes.
To deal with such AMR symbols, we follow previous work~\cite{Bevilacqua_Blloshmi_Navigli_2021} and expand the vocabulary by adding all relations and frames.
In addition, to distinguish between texts and AMR graphs, we add two special tokens, \texttt{<g>} and \texttt{<}$/$\texttt{g>}, to mark the beginning and end of AMR graphs, respectively.

\subsection{Pre-training on AMR graphs}
\label{sec:graphrec}



We introduce two self-supervised training strategies to further pre-train a BART model on AMR graphs. 
As shown in Figure~\ref{fig:graph-masking}(a), the node/edge level denoising strategy encourages the model to capture local knowledge about nodes and edges. 
The graph level denoising strategy (Figure~\ref{fig:graph-masking}(c)) enforces the model to predict a sub-graph, thus facilitating the graph-level learning. 

1) Node/edge level denoising.
We apply a noise function on AMR nodes/edges to construct a noisy input graph.
In particular, the noise function is implemented by masking $15\%$ nodes and $15\%$ edges in each graph.
As shown in Figure~\ref{fig:graph-masking}(a), the node \texttt{[go-01]} and edge \texttt{[:arg0]} are replaced with two \texttt{[mask]} tokens. 

2) Sub-graph level denoising.
This task aims to recover the complete graph when given part of the graph.
We randomly remove a sub-graph\footnote{We define a sub-graph has at least one edge and one node.} from the graph and replace it with a \texttt{[mask]} token (\textit{cf.} Figure~\ref{fig:graph-masking}(c)). 
The masking probability is $0.35$.

\subsection{Unified Pre-training Framework}
\label{sec:dualrec}
\begin{table*}
	\centering
	\small
	\begin{tabular}{llccc}
		\toprule
        \multicolumn{2}{l}{\textbf{Phase}} & \textbf{Task} & \textbf{Input} & \textbf{Output} \\
		\midrule 	
		\multirow{4}{*}{\textbf{(a)}} & \multirow{2}{*}{\textbf{Std. P.T.}} &$\hat{\texttt{t}}$\texttt{2t} & \texttt{<s>} $x_1,..\texttt{[mask]}..,x_n$ \texttt{<}$/$\texttt{s>} & \texttt{<s>} $x_1,x_2,...,x_n$ \texttt{<}$/$\texttt{s>} \\
		& &$\hat{\texttt{g}}$\texttt{2g} &\texttt{<g>} $g_1,..\texttt{[mask]}..,g_m$ \texttt{<}$/$\texttt{g>} & \texttt{<g>} $g_1,g_2,...,g_m$ \texttt{<}$/$\texttt{g>} \\
		\cdashline{2-5}[4pt/2pt]
		&\multirow{2}{*}{\textbf{Std. F.T.}} & \texttt{g2t} & \texttt{<g>} $g_1,g_2,...,g_m$ \texttt{<}$/$\texttt{g>} & \texttt{<s>} $x_1,x_2,...,x_n$ \texttt{<}$/$\texttt{s>} \\
		& & \texttt{t2g} &\texttt{<s>} $x_1,x_2,...,x_n$ \texttt{<}$/$\texttt{s>} & \texttt{<g>} $g_1,g_2,...,g_m$ \texttt{<}$/$\texttt{g>}\\
		\midrule
		\multirow{8}{*}{\textbf{(b)}} & \multirow{6}{*}{\textbf{Unified P.T.}} 
		&$\hat{\texttt{t}}\overline{\texttt{g}}$\texttt{2t} & \texttt{<s>} $x_1,..\texttt{[mask]}..,x_n$ \texttt{<}$/$\texttt{s>} \texttt{<g>} $\texttt{[mask]}$ \texttt{<}$/$\texttt{g>} & \texttt{<s>} $x_1,x_2,...,x_n$ \texttt{<}$/$\texttt{s>} \\
		& &$\overline{\texttt{t}}\hat{\texttt{g}}$\texttt{2g} &\texttt{<s>} $\texttt{[mask]}$ \texttt{<}$/$\texttt{s>}\texttt{<g>} $g_1,..\texttt{[mask]}..,g_m$ \texttt{<}$/$\texttt{g>} & \texttt{<g>} $g_1,g_2,...,g_m$ \texttt{<}$/$\texttt{g>}\\
		& & $\hat{\texttt{t}}$\texttt{g2t} & \texttt{<s>} $x_1,..\texttt{[mask]}..,x_n$ \texttt{<}$/$\texttt{s>} \texttt{<g>} $g_1,g_2,...,g_m$ \texttt{<}$/$\texttt{g>} & \texttt{<s>} $x_1,x_2,...,x_n$ \texttt{<}$/$\texttt{s>} \\
		& & \texttt{t}$\hat{\texttt{g}}$\texttt{2g} & \texttt{<s>} $x_1,x_2,...,x_n$ \texttt{<}$/$\texttt{s>} \texttt{<g>} $g_1,..\texttt{[mask]}..,g_m$ \texttt{<}$/$\texttt{g>} & \texttt{<g>} $g_1,g_2,...,g_m$ \texttt{<}$/$\texttt{g>}\\
		& & $\hat{\texttt{t}}\hat{\texttt{g}}$\texttt{2t} & \texttt{<s>} $x_1,..\texttt{[mask]}..,x_n$ \texttt{<}$/$\texttt{s>} \texttt{<g>} $g_1,..\texttt{[mask]}..,g_m$ \texttt{<}$/$\texttt{g>} & \texttt{<s>} $x_1,x_2,...,x_n$ \texttt{<}$/$\texttt{s>} \\
		& & $\hat{\texttt{t}}\hat{\texttt{g}}$\texttt{2g} & \texttt{<s>} $x_1,..\texttt{[mask]}..,x_n$ \texttt{<}$/$\texttt{s>} \texttt{<g>} $g_1,..\texttt{[mask]}..,g_m$ \texttt{<}$/$\texttt{g>} & \texttt{<g>} $g_1,g_2,...,g_m$ \texttt{<}$/$\texttt{g>}\\
		\cdashline{2-5}[4pt/2pt]
		& \multirow{2}{*}{\textbf{Unified F.T.}} & $\overline{\texttt{t}}\texttt{g}$\texttt{2t} & \texttt{<s>} $\texttt{[mask]}$ \texttt{<}$/$\texttt{s>} \texttt{<g>} $g_1,g_2,...,g_m$ \texttt{<}$/$\texttt{g>} & \texttt{<s>} $x_1,x_2,...,x_n$ \texttt{<}$/$\texttt{s>} \\
		& & $\texttt{t}\overline{\texttt{g}}$\texttt{2g} &\texttt{<s>} $x_1,x_2,...,x_n$ \texttt{<}$/$\texttt{s>} \texttt{<g>} $\texttt{[mask]}$ \texttt{<}$/$\texttt{g>} & \texttt{<g>} $g_1,g_2,...,g_m$ \texttt{<}$/$\texttt{g>}\\
		\bottomrule
	\end{tabular}
	\caption{Different pre-training and fine-tuning strategies.
	P.T. = pre-training, F.T. = fine-tuning. $\texttt{t\\/g}$ denotes the \textit{original} text/graph. $\hat{\texttt{t}} / \hat{\texttt{g}}$ represents a \textit{noisy} text/graph. $\overline{\texttt{t}} / \overline{\texttt{g}}$ means an \textit{empty} text/graph.}
	\label{tab:trainingphase}
\end{table*}
The above standard pre-training and fine-tuning strategies are shown in Table~\ref{tab:trainingphase}(a), 
by using \texttt{<s>} and \texttt{<g>} for differentiating text and graphic information, respectively.
However, the model does not fully learn the interaction between textual and AMR information during pre-training. 
To further address this issue, we consider a unified pre-training framework, which combines text and AMR sequences as input to the denoise auto-encoder. In this way, dynamic masking can be carried out on the text, AMR or both ends, so that the model can learn to make use of one source of information for inferring the other. This can benefit both a parser and a generation model by enforcing the learning of correspondence between text and AMR structures. 

In addition, as shown in Table~\ref{tab:trainingphase}, there is a gap between standard pre-training and fine-tuning for AMR from/to text transduction.
Specifically, the input and output formats are same in the pre-training (\textit{i.e.}, $\hat{\texttt{t}}$\texttt{2t} and $\hat{\texttt{g}}$\texttt{2g}) but different in the fine-tuning (\textit{i.e.}, \texttt{t2g} and \texttt{g2t}). 
This gap restrains models to make the best use of \emph{pre-trained knowledge} in the fine-tuning phase. 
The unified pre-training framework can also benefit task-specific fine-tuning by eliminating the difference of input and output formats between pre-training and fine-tuning. 

Formally, denoting the text and linearized graph sequence as \texttt{t} and \texttt{g}, where $\texttt{t}=\{x_1, x_2,...,x_n\}$ and $\texttt{g}=\{g_1, g_2,...,g_n\}$.
$\hat{\texttt{t}}$ and $\hat{\texttt{g}}$ represent the \textit{noisy} text and graph, respectively, and $\overline{\texttt{t}}$ and $\overline{\texttt{g}}$ refer to the \textit{empty} text and graph, respectively. 
As shown in Table~\ref{tab:trainingphase}(b), we unify the input format for both pre-training and fine-tuning to \texttt{tg}.
For consistency, all input sequences start with a text sequence and end with a graph sequence.


\noindent\textbf{Joint Text and Graph Pre-training.}
We introduce $4$ auxiliary pre-training tasks to encourage information exchanging between graphs and text. 
As shown in Table~\ref{tab:trainingphase}(b), the auxiliary tasks are:

1) Graph augmented text denoising ($\hat{\texttt{t}}$\texttt{g2t}), where an AMR graph is taken as additional input to help masked text reconstruction; 

2) Text augmented graph denoising (\texttt{t}$\hat{\texttt{g}}$\texttt{2g}), where text helps masked graph reconstruction;

3) Noisy graph augmented text denoising ($\hat{\texttt{t}}\hat{\texttt{g}}$\texttt{2t}), where the target text is generated based on a pair of masked text and masked graph;

4) Noisy text augmented graph denoising ($\hat{\texttt{t}}\hat{\texttt{g}}$\texttt{2g}), where a target graph is generated based on a pair of masked text and masked graph.


\noindent\textbf{Dynamic masking rate.} Different from standard masking~\cite{devlin-etal-2019-bert} that uses a static masking rate, we adopt a dynamic masking rate $p$ for task $\hat{\texttt{t}}$\texttt{g2t} and \texttt{t}$\hat{\texttt{g}}$\texttt{2g}. 
Formally, at step $t$, we calculate the masking probability $p$ as:
\begin{equation}
\label{eq:maskrate}
p = 0.1 + 0.75 * t/T,
\end{equation}
where $0.1$ is the initial masking rate, $T$ denotes the total training step. 
$p$ increases as $t$ grows, as $t$ approaches to $T$, the pre-training tasks $\hat{\texttt{t}}$\texttt{g2t} and \texttt{t}$\hat{\texttt{g}}$\texttt{2g} are closer to fine-tuning tasks.

\noindent\textbf{Unified Pre-training and Fine-tuning.} 
In our unified framework, fine-tuning tasks can be viewed as having an \textit{empty} text/graph in the original input, resulting in an input format of $\overline{\texttt{t}}\texttt{g}$\texttt{2t} for AMR-to-text generation and $\texttt{t}\overline{\texttt{g}}$\texttt{2g} for AMR parsing.
In this way, pre-training and fine-tuning tasks share the same input format, thus facilitating knowledge transfer from pre-training to fine-tuning.

\subsection{Training}
To pre-train our model, we optimize the total loss ($\mathcal{L}_{total}$) which is calculated as:
\begin{equation}
\begin{split}
\mathcal{L}_{\hat{\texttt{t}}\texttt{2t}} &=   -\log P(\texttt{t}|\hat{\texttt{t}}, \overline{\texttt{g}}), \\
\mathcal{L}_{\hat{\texttt{g}}\texttt{2g}} &=   -\log P(\texttt{g}|\overline{\texttt{t}}, \hat{\texttt{g}}), \\
\mathcal{L}_{\hat{\texttt{t}}\texttt{g2t}} &=   -\log P(\texttt{t}|\hat{\texttt{t}}, \texttt{g}), \\
\mathcal{L}_{{\texttt{t}}\hat{\texttt{g}} \texttt{2g}} &=   -\log P(\texttt{g}|{\texttt{t}}, \hat{\texttt{g}}), \\
\mathcal{L}_{\hat{\texttt{t}}\hat{\texttt{g}} \texttt{2t}} &=   -\log P(\texttt{t}|\hat{\texttt{t}}, \hat{\texttt{g}}), \\
\mathcal{L}_{\hat{\texttt{t}}\hat{\texttt{g}} \texttt{2g}} &=   -\log P(\texttt{g}|\hat{\texttt{t}}, \hat{\texttt{g}}), \\
\mathcal{L}_{total} &= \mathcal{L}_{\hat{\texttt{t}}\texttt{2t}} + \mathcal{L}_{\hat{\texttt{g}}\texttt{2g}} + \mathcal{L}_{\hat{\texttt{t}}\texttt{g2t}} \\ & +\mathcal{L}_{{\texttt{t}}\hat{\texttt{g}} \texttt{2g}} + \mathcal{L}_{\hat{\texttt{t}}\hat{\texttt{g}} \texttt{2t}} + \mathcal{L}_{\hat{\texttt{t}}\hat{\texttt{g}} \texttt{2g}}, \\
\end{split}
\end{equation}
where $\mathcal{L}_{\hat{\texttt{t}}\texttt{2t}}$ and $\mathcal{L}_{\hat{\texttt{g}}\texttt{2g}}$ are standard pre-training loss on text (Section~\ref{sec:bart}) and graph (Section~\ref{sec:graphrec}), respectively. 
$\mathcal{L}_{\hat{\texttt{t}}\texttt{g2t}}, \mathcal{L}_{{\texttt{t}}\hat{\texttt{g}} \texttt{2g}}, \mathcal{L}_{\hat{\texttt{t}}\hat{\texttt{g}} \texttt{2t}}$ and $\mathcal{L}_{\hat{\texttt{t}}\hat{\texttt{g}} \texttt{2g}}$ denote joint pre-training losses (Section~\ref{sec:dualrec}), respectively.

For fine-tuning, the training objectives are:
\begin{equation}
\begin{split}
\mathcal{L}_{\texttt{amr2text}} &=   -\log P(\texttt{t}|\overline{\texttt{t}},\texttt{g}), \\
\mathcal{L}_{\texttt{text2amr}} &=   -\log P(\texttt{g}|\texttt{t},\overline{\texttt{g}}), \\
\end{split}
\end{equation}
where $\mathcal{L}_{\texttt{amr2text}}$ and $\mathcal{L}_{\texttt{text2amr}}$ are training loss of AMR-to-text generation and AMR parsing, respectively.
\section{Experiments}
We evaluate the effectiveness of our model on five benchmarks and compare the results with state-of-the-art models on AMR parsing and AMR-to-text generation, respectively.
In addition to standard supervised training settings, we evaluate the robustness of our model in a zero-shot domain adaptation setting. 

\subsection{Datasets}
Table~\ref{tab:datasets} shows the statistics of datasets. 
Following~\citet{Bevilacqua_Blloshmi_Navigli_2021}, we use the \textbf{AMR2.0} (LDC2017T10)
and \textbf{AMR3.0} (LDC2020T02).
We also evaluate the model performance on \textbf{New3}, \textit{The Little Prince} (\textbf{TLP}) and \textit{Bio AMR} (\textbf{Bio}) corpora.
For pre-training, we additionally use 200k silver data parsed by SPRING~\cite{Bevilacqua_Blloshmi_Navigli_2021}. 
These data are randomly selected from Gigaword (LDC2011T07)
corpus, which shares the same textual source with AMR data.\footnote{\url{https://catalog.ldc.upenn.edu}.}

\subsection{Settings}
We follow~\citet{Bevilacqua_Blloshmi_Navigli_2021} in pre-processing and post-processing AMR graphs, except for omitting the recategorization step which does not consistently improve model performance in our preliminary experiments.
Our model is built based on a vanilla BART\footnote{\url{https://github.com/huggingface}.}.
The best model and hyper-parameters are selected by performance on the validation set.
The detailed hyper-parameters are given in Appendix~\ref{appendix:para}.

\begin{table}
	\centering
	\small
	\begin{tabular}{lccccc}
		\toprule
        \textbf{Datasets} & AMR2.0 & AMR3.0  & New3 & TLP &Bio\\
		\midrule 	
		Train & 36521 & 55635 & - & - & - \\
		Valid & 1368 & 1722 &- & - & - \\
		Test & 1371 & 1898 & 527 & 1562 & 500 \\
	   \bottomrule
	\end{tabular}
	\caption{Benchmark AMR datasets.}
	\label{tab:datasets}
\end{table}

\noindent\textbf{Metrics.}
Following~\citet{Bevilacqua_Blloshmi_Navigli_2021}, we evaluate on the AMR parsing benchmarks by using Smatch~\cite{cai-knight-2013-smatch} and other fine-grained metrics.\footnote{Please refer to Appendix~\ref{appendix:metrix} for more details.}
Regarding AMR-to-text, we use three common Natural Language Generation measures, including BLEU ~\cite{papineni-etal-2002-bleu}, CHRF++ ~\cite{popovic-2017-chrf} and METEOR~\cite{banerjee-lavie-2005-meteor}, tokenizing with the script provided with JAMR~\cite{flanigan-etal-2014-discriminative}.

\begin{table}[!t]
	\centering
	\small
	\begin{tabular}{lccc}
		\toprule
		\textbf{Setting} &Smatch &BLEU &Avg\\
		\midrule
		BART-base  &82.7 &42.5 &62.6 \\
        + $\hat{\texttt{t}}\overline{\texttt{g}}$\texttt{2t} &82.9 &42.9 &62.9 \\
		+ $\overline{\texttt{t}}\hat{\texttt{g}}$\texttt{2g} &83.1 &42.6 &62.9 \\
		+ $\hat{\texttt{t}}\overline{\texttt{g}}$\texttt{2t}, $\overline{\texttt{t}}\hat{\texttt{g}}$\texttt{2g} &83.1 &42.8 &63.0 \\
		+ $\hat{\texttt{t}}\overline{\texttt{g}}$\texttt{2t}, $\overline{\texttt{t}}\hat{\texttt{g}}$\texttt{2g}, ${\texttt{t}}\hat{\texttt{g}}$\texttt{2g} &83.4 & 42.8 & 63.1 \\
		+ $\hat{\texttt{t}}\overline{\texttt{g}}$\texttt{2t}, $\overline{\texttt{t}}\hat{\texttt{g}}$\texttt{2g}, $\hat{\texttt{t}}{\texttt{g}}$\texttt{2t} &83.1 &45.3 &63.2 \\
		+ $\hat{\texttt{t}}\overline{\texttt{g}}$\texttt{2t}, $\overline{\texttt{t}}\hat{\texttt{g}}$\texttt{2g}, ${\texttt{t}}\hat{\texttt{g}}$\texttt{2g}, $\hat{\texttt{t}}{\texttt{g}}$\texttt{2t} &83.3 &45.0 &63.2 \\
		+ $\hat{\texttt{t}}\overline{\texttt{g}}$\texttt{2t}, $\overline{\texttt{t}}\hat{\texttt{g}}$\texttt{2g},
		$\hat{\texttt{t}}\hat{\texttt{g}}$\texttt{2g}
		&83.2 &43.0 &63.1 \\
		+ $\hat{\texttt{t}}\overline{\texttt{g}}$\texttt{2t}, $\overline{\texttt{t}}\hat{\texttt{g}}$\texttt{2g},
		$\hat{\texttt{t}}\hat{\texttt{g}}$\texttt{2t}
		&83.1 &44.2 &63.7 \\
		+ $\hat{\texttt{t}}\overline{\texttt{g}}$\texttt{2t}, $\overline{\texttt{t}}\hat{\texttt{g}}$\texttt{2g},
		$\hat{\texttt{t}}\hat{\texttt{g}}$\texttt{2g}, 
		$\hat{\texttt{t}}\hat{\texttt{g}}$\texttt{2t} & 83.2 &44.0 &63.6 \\
		+ \texttt{ALL} &83.6 &45.6 &64.1 \\
		\bottomrule
	\end{tabular}
	\caption{AMP parsing (Smatch) and AMR-to-text generation (BLEU) performance on valid set of AMR2.0.}
	\label{tab:dev-exp}
\end{table}

\subsection{Compared Models}
For \textbf{AMR parsing}, we consider following systems for comparison: 1) \citeauthor{TitovL18} (2018; LyuT), a neural parser trained by jointly modeling alignments, concepts and relations;
2) \citeauthor{zhang-etal-2019-broad} (2019b; Zhang+), a seq2seq approach that incrementally builds up an AMR via predicting semantic relations;
3) \citeauthor{zhou2020amr} (2020; Zhou+), an aligner-free parser enhanced by explicit dependency and latent structures;
4) \citeauthor{cai-lam-2020-amr} (2020a; CaiL), a graph-based parser that enhances incremental sequence-to-graph model with a graph-sequence iterative inference mechanism;
5) \citeauthor{Bevilacqua_Blloshmi_Navigli_2021} (2021; Bevilacqua+), a fine-tuned BART model that predicts a linearized AMR graph.

For \textbf{AMR-to-text generation}, the compared models are: 
1) \citeauthor{zhu2019modeling} (2019; Zhu+), a Transformer-based model that enhances self-attention with graph relations; 
2) \citeauthor{zhang2020lightweight} (2020; Zhang+), a graph-to-sequence model which uses a dynamic graph convolutional networks for better graph modeling.
3) \citeauthor{bai-etal-2020-online} (2020; Bai+), a graph encoder~\cite{zhu2019modeling} with a structural decoder that jointly predicts the target text and the input structure;
4) \citeauthor{mager-etal-2020-gpt} (2020; Mager+), a fine-tuned GPT that predicts text based on a PENMAN linearized AMR graph;
5) \citeauthor{Bevilacqua_Blloshmi_Navigli_2021} (2021; Bevilacqua+), a fine-tuned BART that predicts text based on a DFS linearized AMR graph;
6) \citeauthor{ribeiro-etal-2021-investigating} (2021; Ribeiro+), a fine-tuned BART based on a PENMAN linearized AMR graph.
For a fair comparison, we leave out models based on T5~\cite{ribeiro-etal-2021-investigating,Ribeiro2021StructuralAI}, which has about two times more parameters than BART.

\subsection{Development Experiments}
\label{sec:dev-exp}
Table~\ref{tab:dev-exp} shows results on the validation set of AMR2.0 under different model settings, where we take a fine-tuned BART-based model~\cite{Bevilacqua_Blloshmi_Navigli_2021} as our baseline.

We first study the effectiveness of pre-training only on text and graphs.
As shown in Table~\ref{tab:dev-exp}, both pre-training on the text ($\hat{\texttt{t}}\overline{\texttt{g}}$\texttt{2t}) and graph ($\overline{\texttt{t}}\hat{\texttt{g}}$\texttt{2g}) leads to better results, and combining them can give better results on both tasks.
Also, adding joint pre-training tasks improves the performance. 
In particular, ${\texttt{t}}\hat{\texttt{g}}$\texttt{2g} gives a Smatch improvement of $0.7$ for AMR paring, and $\hat{\texttt{t}}{\texttt{g}}$\texttt{2t} reaches a BLEU of $45.3$ for AMR-to-text generation, which is $2.8$ points higher than baseline. 
Adding $\hat{\texttt{t}}\hat{\texttt{g}}$\texttt{2g} gives a Smatch of $83.2$ for AMR parsing, and $\hat{\texttt{t}}\hat{\texttt{g}}$\texttt{2t} improves the baseline by $1.7$ BLEU points for generation.
By combining ${\texttt{t}}\hat{\texttt{g}}$\texttt{2g} and $\hat{\texttt{t}}{\texttt{g}}$\texttt{2t}, the performance increase by $0.6$ and $2.5$ points on AMR parsing and AMR-to-text generation, respectively.
Similar trend can be observed by combining $\hat{\texttt{t}}\hat{\texttt{g}}$\texttt{2g} and $\hat{\texttt{t}}\hat{\texttt{g}}$\texttt{2t}.
Finally, using all $6$ pre-training tasks, our model reach a result of $83.6$ Smatch and $45.6$ BLEU, respectively.

\begin{table}
	\centering
	\small
	\begin{tabular}{lcc}
		\toprule
        \textbf{Setting} & AMR parsing & AMR-to-text \\
		\midrule 	
		Full Model & 83.6 & 45.6  \\
		\quad - Node/edge masking & 83.4 & 45.1 \\
		\quad - Sub-graph masking & 83.1 & 44.7 \\
	\bottomrule
	\end{tabular}
	\caption{Impact of two masking strategies.}
	\label{tab:ablation}
\end{table}

We also study the impact of two graph self-supervised training strategies.
In particular, we evaluate the performance after removing the node/edge or the sub-graph masking task independently. 
As shown in Table~\ref{tab:ablation}, 
the performance decreases on both AMR parsing and AMR-to-text generation tasks without the node/edge level masking strategy.
The performance drop is larger when removing the sub-graph masking task, with a margin of $0.5$ Smatch and $0.9$ BLEU, respectively.

\begin{figure}[!t]
	\setlength{\belowcaptionskip}{-0.1cm}
	\centering 
	\subfigure[]{\includegraphics[width=0.45\hsize]{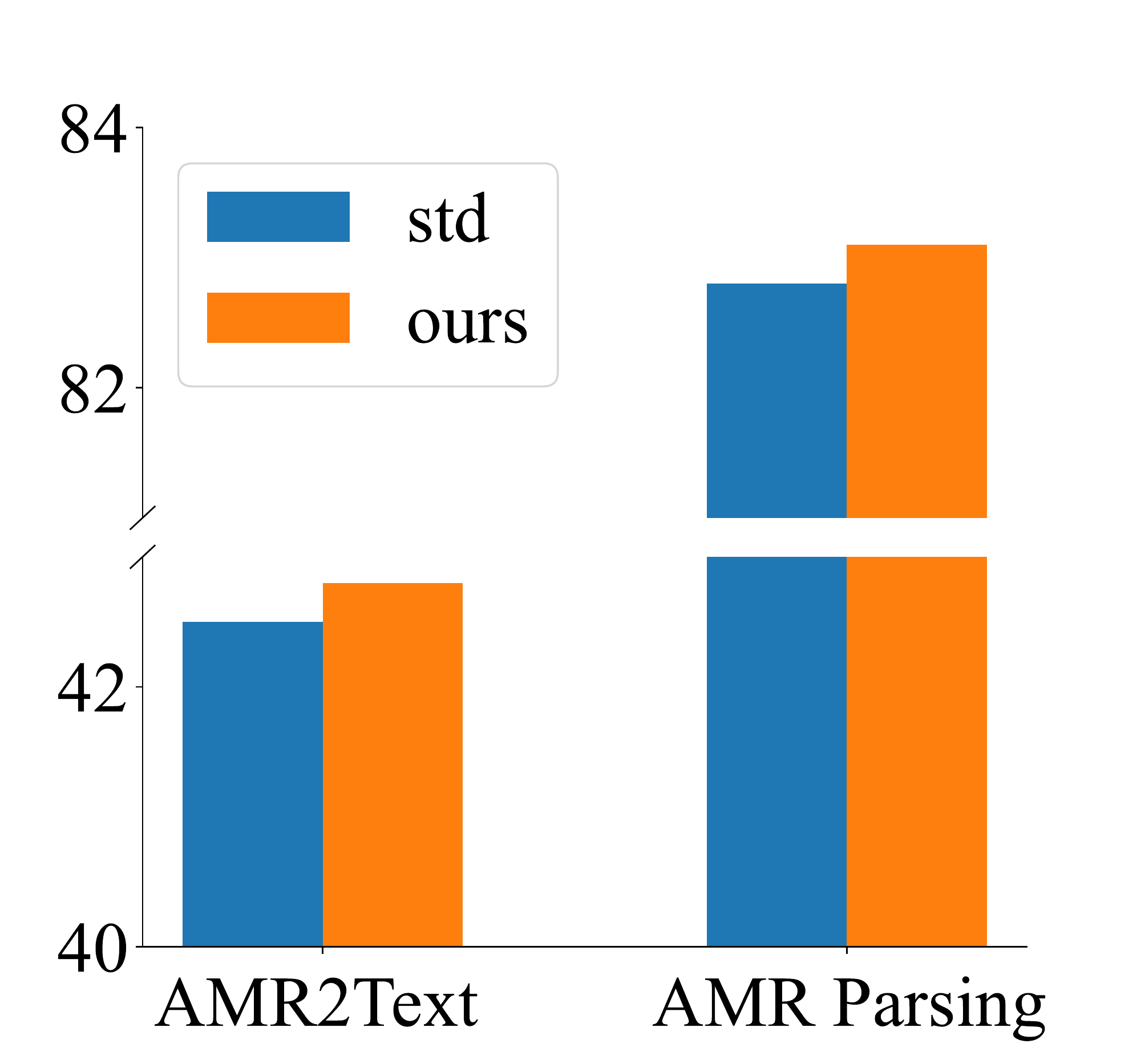}\label{fig:unified}} \hspace{0.10in}
	\subfigure[]{\includegraphics[width=0.43\hsize]{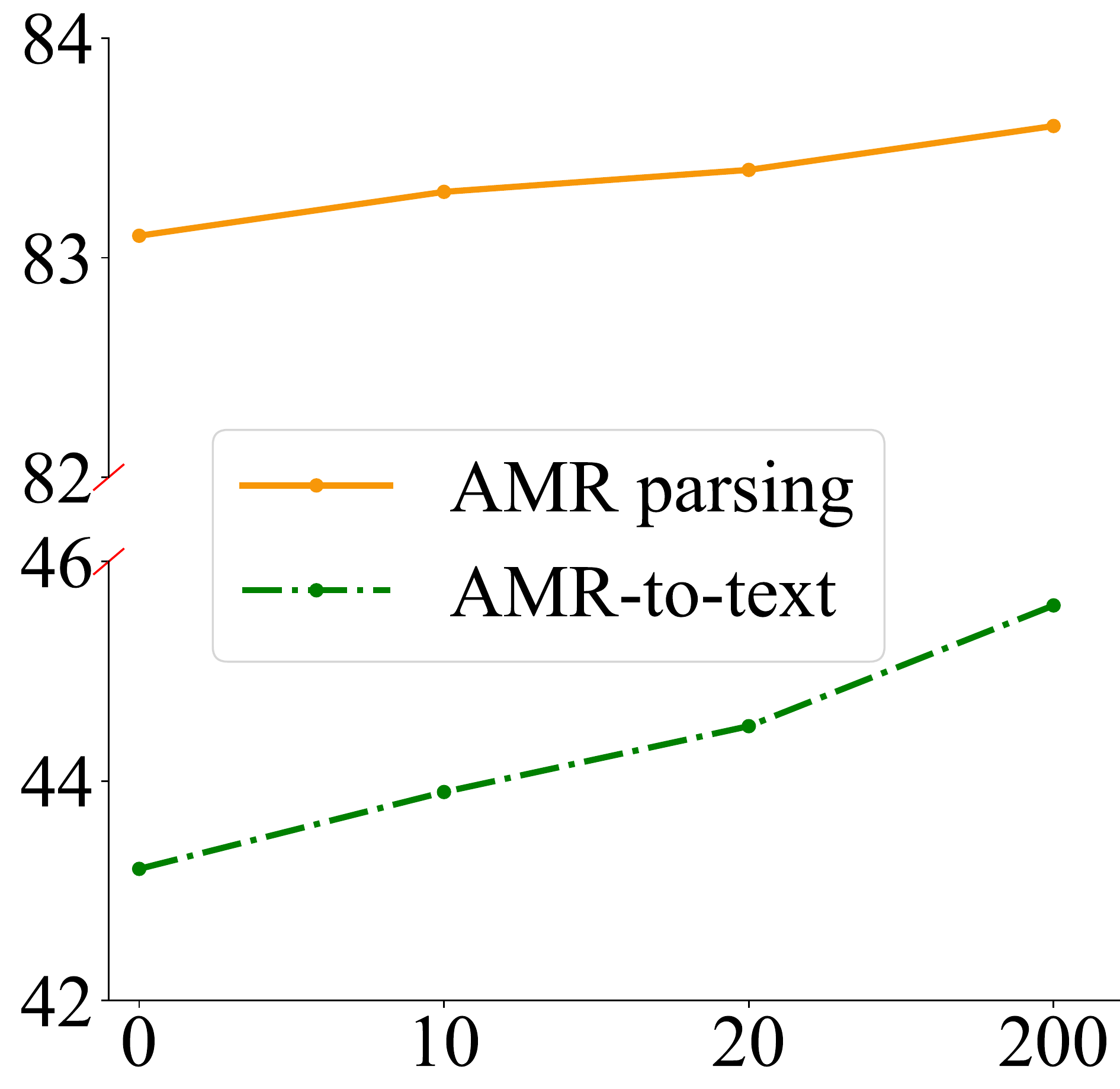}\label{fig:silver}} 
	\caption{Development results: (a) comparison of standard pre-training and fine-tuning phase (std) and our unified frameworks; (b) impact of silver data.}
	\label{fig:dev}
\end{figure}

Figure~\ref{fig:unified} compares the performance of standard pre-training (\texttt{$\hat{\texttt{t}}$\texttt{2t}}, $\hat{\texttt{g}}$\texttt{2g}) and fine-tuning (\texttt{t2g}, \texttt{g2t}) with our unified framework. 
The unified framework gives better results than standard versions on both tasks. 
This confirms our assumption that our unified framework is helpful for reducing the gap between pre-training and fine-tuning.
Besides, we find that by unifying pre-training and fine-tuning formats, our model converges faster than the baseline during fine-tuning (\emph{cf.} Appendix~\ref{appendix:effect}).

Figure~\ref{fig:silver} shows the model performance regarding different scales of silver data. 
Even without silver data, the performance of our model is better than the baseline, indicating that graph pre-training is beneficial for downstream tasks when using various auxiliary tasks.
When silver data are available, the performance of both AMR parsing and AMR-to-text generation tasks increases as the scale of silver data increases, with a margin of $2$ BLEU score.
We also \emph{fine-tune} a BART model on silver data under our unified framework (i.e., $\overline{\texttt{t}}\texttt{g}$\texttt{2t} and $\texttt{t}\overline{\texttt{g}}$\texttt{2g}), and find that our dual graph and text denoising tasks are more useful (\textit{cf.} Appendix~\ref{appendix:denoising} for more analysis and discussion).



\begin{table*}[!t]
	\centering
	\small
	\begin{tabular}{l|c|cccccccc}
		\toprule
        \textbf{Model} & \textbf{Smatch} & \textbf{Unlab.}  & \textbf{NoWSD} & \textbf{Con.} &\textbf{Wiki.} & \textbf{NER} & \textbf{Reent.} & \textbf{Neg.} & \textbf{SRL}\\
		\midrule
		\textbf{AMR2.0} & & & & & & & & &\\
		LyuT (2018) &74.4 &77.1 &75.5 &85.9 &75.7 &86.0 &52.3 &58.4 &69.8 \\
		Zhang+ (2019b)$^\dag$ &77.0 &80.0 &78.0 &86.0 &86.0 &79.0 &61.0 &77.0 &71.0 \\
		Zhou+ (2020)$^\dag$ &77.5 &80.4 &78.2 &85.9 &86.5 &78.8 &61.1 &76.1 &71.0 \\
		CaiL (2020a)$^\dag$ &80.2 &82.8 &80.0 &88.1 &86.3 &81.1 &64.6 &78.9 &74.2 \\
		Xu+ (2020)$^\dag$ &80.2 &83.7 &80.8 &87.4 &75.1 &85.4 &66.5 &71.5 &78.9 \\
		Bevilacqua+ (2021, base)$^\dag$ &82.7 &85.1 &83.3 &89.7 &82.2 &90.0 &70.8 &72.0 &79.1 \\
		Bevilacqua+ (2021, large)$^\dag$ &84.5 &86.7 &84.9 &89.6 &\textbf{87.3} &83.7 &72.3 &\textbf{79.9} &79.7 \\
		Bevilacqua+ (2021, large)$^{\dag s}$ &84.3 &86.7 &84.8 &90.8 &83.1 &90.5 &72.4 &73.6 &80.5 \\
		\rowcolor{mygray}
		Ours (base)$^\dag$ & 83.6 &86.7 &84.0 &90.2 &78.6 &90.0 &71.3 &73.7 &79.5 \\
		\rowcolor{mygray}
		Ours (large)$^\dag$ &\textbf{85.4} &\textbf{88.3} &\textbf{85.8} &\textbf{91.2} &81.4 &\textbf{91.5} &\textbf{73.5} &74.0 &\textbf{81.5} \\
	   \midrule
		\textbf{AMR3.0} & & & & & & & & & \\
		Bevilacqua+ (2021, large)$^\dag$ &83.0 &85.4 &83.5 &89.8 &\textbf{82.7} &87.2 &70.4 &\textbf{73.0} &78.9 \\
		Bevilacqua+ (2021, large)$^{\dag s}$ &83.0 &85.4 &83.5 &89.5 &81.2 &87.1 &71.3 &71.7 &79.1 \\
		\rowcolor{mygray}
		Ours (base)$^\dag$ &82.5 &85.7 &82.9 &89.4 &76.1 &86.8 &69.9 &70.3 &78.2 \\
		\rowcolor{mygray}
		Ours (large)$^\dag$ & \textbf{84.2} &\textbf{87.1} &\textbf{84.6} &\textbf{90.2} &78.9 &\textbf{88.5} &\textbf{72.4} &72.1 &\textbf{80.3} \\
		\bottomrule
	\end{tabular}
	\caption{AMR parsing results on AMR2.0 and AMR3.0. $s$ means the model uses 200k silver data for fine-tuning. $\dag$ means the model is based on pre-trained models. The best result within each row block is shown in bold.}
	\label{tab:main-parsing}
\end{table*}

\subsection{Main Results}
\label{sec:main-res}
\noindent\textbf{AMR parsing}. Table~\ref{tab:main-parsing} lists the result of different models on AMR2.0 and AMR3.0.
Among previous works, Bevilacqua+ (2021, large) achieves the best results, consistently outperforming other systems. 
Compared with the system of~\citet{Bevilacqua_Blloshmi_Navigli_2021}, our model obtains significantly ($p$<0.01) better Smatch scores in both \textit{base} and \textit{large} settings on both datasets.
In particular, our \textit{base} model outperforms the Bevilacqua+ (2021, base) by $0.9$ Smatch point on AMR2.0, and our \textit{large} model obtains a Smatch of $85.4$ and $84.2$ on AMR2.0 and AMR3.0, respectively. To our knowledge, these are the best-reported results, showing the effectiveness of our method.


Besides, Bevilacqua+ (2021, large)$^s$ uses silver data for fine-tuning, yet does not lead to consistent improvement over Bevilacqua+ (2021, large). 
In contrast, our large model gives $1.1$ and $1.2$ higher Smatch than Bevilacqua+ (2021, large)$^s$ on AMR2.0 and AMR3.0, respectively.
This indicates that our pre-training framework is a better way than fine-tuning to make use of silver data.
The main reason is that our models are pre-trained using a denoising auto-encoding manner, which is less sensitive to silver (or noisy) data than fine-tuning.
We also find that further fine-tuning our models on silver data (same with pre-training) cannot bring improvement (\textit{cf.} Appendix~\ref{appendix:silverdata}).


\noindent\textbf{AMR-to-text generation}. We report the results of different systems on AMR2.0 and AMR3.0 in Table~\ref{tab:main-amr2text}, respectively. With the help of BART, Ribeiro+ (2021) and Bevilacqua+ (2021, large) obtain significantly better results than previous graph-to-sequence and GPT-based models.
Compared with Bevilacqua+ (2021), our models (\textit{base} and \textit{large}) give significantly ($p$<0.001) better results in terms of all evaluation metrics.
In particular, our \textit{base} model achieves comparable or better performance than Bevilacqua+ (2021, large).
Compared with Bevilacqua+ (2021, large)$^s$, our large model improves the performance by $3.9$ and $2.7$ points on AMR2.0 and AMR3.0, respectively.
Similar with AMR parsing, we  observe that when fine-tuning our models on silver data cannot bring improvement for AMR-to-text generation task (Table~\ref{tab:main-amr2text} and Appendix~\ref{appendix:silverdata}).


\noindent\textbf{Out-of-distribution evaluation} 
We use the model trained on AMR2.0 to get predictions on out-of-domain test sets. 
Table~\ref{tab:ood} shows the results on AMR parsing and AMR-to-text generation tasks.
Similar to in-domain experiments, our models achieve better results than existing methods.
In particular, our \textit{base} model can give comparable performance than Bevilacqua+ (2021, large), and our \textit{large} model obtains the best-reported results. 
This indicates that our model is more robust to new domains, thanks to joint graph and text pre-training.
Regarding different domains, our method achieves bigger improvements on New3 than the other two domains.
This is intuitive, as pre-training strengthens the model representation power on the domain of graph pre-training data, and New3 is closer to it than other two datasets.
\begin{table}[!t]
	\centering
	\small
	\begin{tabular}{lccc}
		\toprule
		\textbf{Model} &\textbf{BLEU} &\textbf{CH.} &\textbf{MET.} \\
		\midrule
		\textbf{AMR2.0} &&&\\
		Zhu+ (2019) &31.8 &64.1 &36.4\\
		Zhang+ (2020) &33.6 &63.2 &37.5 \\
		Bai+ (2020) &34.2 &65.7 &38.2\\
		Mager+ (2020)$^\dag$ &33.0 &63.9 &37.7\\
		Ribeiro+ (2021)$^{\dag\ddagger}$ &45.9 &- &{41.2}\\
		Bevilacqua+ (2021, base)$^\dag$ &42.7 &72.2 &40.7\\
		Bevilacqua+ (2021, large)$^\dag$ &45.3 &73.5 &41.0\\
		Bevilacqua+ (2021, large)$^{s\dag}$ &45.9 &74.2 &41.8\\
		\rowcolor{mygray}
		Ours (base)$^\dag$ &46.6 &74.6 &41.4 \\
		\rowcolor{mygray}
		Ours (large)$^\dag$ &\textbf{49.8} &\textbf{76.2} &\textbf{42.6} \\
		\midrule
		\textbf{AMR3.0} &&&\\
		Zhang+ (2020) &34.3 &63.7 &38.2 \\
		Bevilacqua+ (2021, large)$^\dag$ &44.9 &72.9 &40.6\\
		Bevilacqua+ (2021, large)$^{s\dag}$ &46.5 &73.9 &41.7\\
		\rowcolor{mygray}
		Ours (base)$^\dag$ &45.9 &73.8 &40.8\\
		\rowcolor{mygray}
		Ours (large)$^\dag$ &\textbf{49.2} &\textbf{76.1} &\textbf{42.3}\\
		\bottomrule
	\end{tabular}
	\caption{AMR-to-text results on AMR2.0 and AMR3.0. CH.=CHRF++. MET.=METEOR. $s$ means the model uses 200k silver data for fine-tuning. Models marked with $\dag$ are based on PLMs. The best result within each row block is shown in bold. $^\ddagger$For fair comparison, we report results of tokenized output of Ribeiro+ (2021).}
	\label{tab:main-amr2text}
\end{table}

In addition, Bevilacqua+ (2021, large)$^s$ gives lower results than Bevilacqua+ (2021, large) in New3 (both tasks) and TLP (only AMR-to-text generation). 
In contrast, our model gives consistent improvements on all $3$ domains. 
This can be because fine-tuning leads to catastrophic forgetting of distributional knowledge~\cite{Kirkpatrick2017OvercomingCF}.

\begin{table}[t]
	\centering
	\small
	\begin{tabular}{lccc}
		\toprule
		\textbf{Model}& \textbf{New3} &\textbf{TLP} &\textbf{Bio} \\
		\midrule
		\textbf{AMR Parsing} & & & \\
		Bevilacqua+ (2021, large) &73.7 &77.3 &59.7 \\
		Bevilacqua+ (2021, large)$^s$ &71.8 &77.5 &59.5 \\
		\rowcolor{mygray}
		Ours (base) &74.4 &77.8 &58.8 \\
		\rowcolor{mygray}
		Ours (large) &\textbf{76.9} &\textbf{79.8} &\textbf{63.2} \\
		\midrule
		\textbf{AMR-to-Text} & & & \\
		Bevilacqua+ (2021, large) &38.8 &25.4 &18.7 \\
		Bevilacqua+ (2021, large)$^s$ &38.2 &25.1 &19.4 \\
		\rowcolor{mygray}
		Ours (base) &41.0 &26.4 &16.9 \\
		\rowcolor{mygray}
		Ours (large) &\textbf{44.8} &\textbf{29.1} &\textbf{20.7} \\
		\bottomrule
	\end{tabular}
	\caption{Out of distribution performance on AMR parsing (Smatch) and AMR-to-text (BLEU).}
	\label{tab:ood}
\end{table}

\subsection{Impact of Graph}

Table~\ref{tab:graph-analysis} shows the effects of the graph size, graph diameter and reentrancies on the performance. 
We split the test set of AMR2.0 into different groups and report the performance improvement over the baseline model ~\cite{Bevilacqua_Blloshmi_Navigli_2021}. 
All models are trained on AMR2.0.
We first consider graph size, which records the number of nodes in an AMR graph.
Our model consistently outperforms the baseline model on both tasks, with the performance gap growing on larger graphs. 
This indicates that our system is more powerful in dealing with larger graphs.
The main reason is that our joint text and graph pre-training mechanism enhances the model with the ability to capture word or span level correlation between text and graph, which is helpful for dealing with long sequence and large graphs.

The graph depth is defined as the longest distance between the AMR node and root node. 
A graph with deeper depth has more long-range dependencies.
For AMR parsing, our model gives a better Smatch than the baseline model on the first two groups of graphs, and a comparable score on graphs with a depth bigger than $6$.
For AMR-to-text generation, our model consistently improves over the baseline model on all graphs, and the improvements are bigger on deeper graphs.
This shows that our model is better for learning more complex graphs.
It can be that our graph masking strategies train the model to learn the relationships between a sub-graph and the remaining graph context, making it easier to understand deep graphs.

\begin{table}
	\centering
	\small
	\begin{tabular}{lccc}
		\toprule
		\textbf{Graph Size} &1-10 (522) &11-20 (556) &>20 (293) \\
		AMR parsing &+0.3 &+1.0 &+0.8 \\
		AMR-to-text &+0.9 &+3.2 &+2.1 \\
		\midrule
		\textbf{Graph Depth} &1-3 (422) &4-6 (667) &>6 (282) \\
		AMR parsing &+0.8 &+0.9 &0.0 \\
		AMR-to-text &+1.2 &+2.3 &+2.8 \\
		\midrule
		\textbf{Reentrancies} &0 (622) &1-3 (712) &>3 (37) \\
		AMR parsing &+1.1 &+0.6 &0.0 \\
		AMR-to-text &+2.0 &+2.7 &+0.4 \\
		\bottomrule
	\end{tabular}
	\caption{Performance improvements on AMR parsing (Smatch) and AMR-to-text (BLEU).}
	\label{tab:graph-analysis}
\end{table}

Reentrancy is the number of nodes that has multiple parents. 
Reentrancies pose difficulties to both AMR parsing and AMR-to-text tasks~\cite{damonte-cohen-2019-structural,szubert-etal-2020-role}.
The more reentrancies, the harder the graph is to be understood.
Our method gives significantly ($p$<0.01) better results on both tasks when the input graphs have less than $4$ reentrancies.
For graphs with more than $3$ reentrancies, the proposed model is $0.4$ better on AMR-to-text generation task and comparable than the baseline model on AMR parsing task.
This means that our system has an overall better ability on learning reentrancies.

\begin{table}[!t]
    \centering
    \small
    \begin{tabular}{l}
        \toprule
        \textbf{Text\#1:} It's getting \textcolor{red}{hard} to keep strong and keep \\ \quad\quad carrying on with life. \\
        \midrule
        \textbf{Gold:}  \\
        \quad (g / get-03      \\
        \quad\quad:ARG1 (a / and \\
            \quad\quad\quad :op1 (k / keep-02 \\
            \quad\quad\quad\quad      :ARG1 (s / strong-02)) \\
            \quad\quad\quad:op2 (k2 / keep-02 \\
            \quad\quad\quad\quad :ARG1 (c /  carry-on-02 \\
            \quad\quad\quad\quad\quad:ARG1 (l / live-01)))) \\
      \quad\quad:ARG2 (h / \textcolor{red}{hard-02}))\\
        \midrule
        \textbf{Baseline:} \\
    \quad (z0 / get-03 \\
    \quad\quad:ARG1 (z1 / and \\
    \quad\quad\quad :op1 (z2 / keep-02 \\
    \quad\quad\quad\quad:ARG1 (z3 / strong-02)) \\
    \quad\quad\quad :op2 (z4 / carry-on-02 \\
    \quad\quad\quad\quad:ARG1 (z5 / life)))) \\
    \midrule
    \textbf{Ours:} \\
    \quad (z0 / get-03  \\
    \quad\quad :ARG1 (z1 / and \\
    \quad\quad\quad:op1 (z2 / keep-02 \\
    \quad\quad\quad\quad:ARG1 (z3 / strong-02)) \\
    \quad\quad\quad:op2 (z4 / keep-02 \\
    \quad\quad\quad\quad:ARG1 (z5 / carry-on-02 \\
    \quad\quad\quad\quad\quad:ARG1 (z6 / life)))) \\
    \quad\quad :ARG2 (z7 / \textcolor{red}{hard-02} \\
    \quad\quad\quad :ARG1 z1)) \\
    \midrule
    \midrule
    \textbf{Text\#2:} \underline{Self harming} is addictive, but you can\\ \quad\quad overcome it. \\
    \midrule
    \textbf{Gold:} \\
    \quad (c / contrast-01   \\
        \quad\quad :ARG1 (a / addictive-02 \\
            \quad\quad\quad :ARG0 (h / harm-01 \\
            \quad\quad\quad\quad :ARG1 (s / self)))\\
      \quad\quad :ARG2 (p / possible-01 \\
            \quad\quad\quad :ARG1 (o / overcome-01 \\
            \quad\quad\quad\quad :ARG0 (y / you) \\
            \quad\quad\quad\quad  :ARG1 h))) \\
    \midrule\textbf{Baseline:} \\
    \quad (z0 / addictive-02  \\
        \quad\quad \underline{:ARG0 (z1 / harm-01} \\
            \quad\quad\quad \underline{:ARG1 z1)}\\
      \quad\quad :concession-of (z2 / possible-01 \\
            \quad\quad\quad :ARG1 (z3 / overcome-01 \\
            \quad\quad\quad\quad :ARG0 (z4 / you) \\
            \quad\quad\quad\quad  :ARG1 z1))) \\
    \midrule\textbf{Ours:} \\
    \quad (z0 / contrast-01   \\
        \quad\quad :ARG1 (z1 / addictive-02 \\
            \quad\quad\quad :ARG0 (z2 / harm-01 \\
            \quad\quad\quad\quad :ARG1 (z3 / self)))\\
      \quad\quad :ARG2 (z4 / possible-01 \\
            \quad\quad\quad :ARG1 (z5 / overcome-01 \\
            \quad\quad\quad\quad :ARG0 (z6 / you) \\
            \quad\quad\quad\quad  :ARG1 z1))) \\
    \bottomrule
    \end{tabular}
    \caption{Two AMR parsing cases. Given a text input, we present the gold AMR graph and two model outputs, parsed by the baseline and our model, respectively.}
    \label{tab:case-parsing-single}
\end{table}

\subsection{Case study}

Table~\ref{tab:case-parsing-single} presents two cases of AMR parsing, with the model outputs generated by our model and the baseline model, and the gold output given the same input sentence. 
As shown in the first case, the baseline model omits the semantic unit ``\textit{hard}'', thus generates an incomplete AMR graph of a different meaning compared with the input sentence.
In contrast, our system preserves the concept ``\textit{hard}'' and transfers the semantic relations correctly, thanks to the modeling of correspondence between text and graph during pre-training.
In the second case, the baseline output includes a \textbf{cyclic} sub-graph (\textit{i.e.}, \texttt{( z1 harm-01 :ARG1 z1 )}), which is contrary to the grammar that \textit{AMRs should be acyclic}. 
Our system gives a valid AMR graph which is semantically similar with gold graph. 

Table~\ref{tab:case-generation-single} lists two AMR graphs and model outputs of our AMR-to-text model and the baseline model.
In the first case, although the baseline generates a fluent sentence, it ignores the concept ``\textit{have-purpose-91}'', resulting in that the generated sentence is of a different meaning compared with the input graph.
In the second AMR graph, ``\textit{before}'' modifies the phrase ``\textit{won many championships}''. 
However, in the baseline output, ``\textit{before}'' is used to modify the phrase ``\textit{participating in international competitions}''.
Compared with the baseline, our system recovers all concepts and maps the modification relationship from the AMR graph to text correctly.
This indicates that our model generates more faithful sentences than the baseline.
\begin{table}[!t]
    \centering
    \small
    \begin{tabular}{l}
        \toprule
        \textbf{AMR\#1:} (h / \textcolor{red}{have-purpose-91} \\
        \quad\quad :ARG1 (t / thing \\
        \quad\quad\quad    :ARG1-of (e / expend-01 \\
        \quad\quad\quad\quad :ARG2 (t2 / transport-01))) \\
        \quad\quad :ARG2 (a / amr-unknown)) \\
        \midrule
        \textbf{Gold:} What is the \textcolor{red}{purpose} of transportation-related \\ \quad\quad expenditures?  \\
        \textbf{Baseline:}  What are the transportation  expenses?\\
        \textbf{Ours:} What is the \textcolor{red}{purpose} of
        transportation expenses?\\
        \midrule
        \midrule
        \textbf{AMR\#2:} \\
        \quad (w / win-01 \\
        \quad\quad :ARG0 (p2 / person \\ 
        \quad\quad\quad:wiki - \\ 
        \quad\quad\quad:name (n / name \\
        \quad\quad\quad\quad:op1 "Fengzhu" \\
        \quad\quad\quad\quad:op2 "Xu")) \\
        \quad\quad :ARG1 (c / championship-02 \\
        \quad\quad\quad :ARG0 p2 \\
        \quad\quad\quad :quant (m / many)) \\
        \quad\quad :time (b / \textcolor{red}{before}) \\
        \quad\quad :part-of (c2 / compete-01 \\
        \quad\quad\quad :mod (i / international))) \\
        \midrule
        \textbf{Gold:} Fengzhu Xu has won many championships in \\ \quad\quad international competitions \textcolor{red}{before}. \\
        \midrule
        \textbf{Baseline:} Fengzhu Xu won many championships \\ \quad\quad  \textcolor{blue}{before} participating in international competitions. \\
        \midrule
        \textbf{Ours:} Fengzhu Xu has won many championships in \\ \quad\quad international competitions \textcolor{red}{before}. \\
    \bottomrule
    \end{tabular}
    \caption{Two AMR-to-text generation cases. Given an AMR graph, we present the gold text and two generated outputs, given by baseline and our model, respectively.}
    \label{tab:case-generation-single}
\end{table}

\section{Conclusion}
We investigated graph pre-training as a complement to text pre-training for AMR parsing and AMR-to-text generation tasks, using a novel unified framework with dual graph and text denoising.
We find that graph pre-training is highly effective for both AMR parsing and AMR -to-text generation, and is a more effective way of making use of silver data compared with fine-tuning. 
Our methods give the best results on multiple benchmarks for both tasks.

\section*{Acknowledgments}
Yue Zhang is the corresponding author. 
We would like to thank anonymous reviewers for their insightful comments.
This work is supported by the National Natural Science Foundation of China under grant No.61976180 and the Tencent AI Lab Rhino-Bird Focused Research Program.

\bibliography{custom}
\bibliographystyle{acl_natbib}

\clearpage
\appendix
\label{sec:appendix}

\section{Model Hyper-Parameters}\label{appendix:para}
\begin{table}[!t]
    \centering
    \small
    \begin{tabular}{l|c}
        \toprule
        \textbf{Param. Name} & \textbf{Value} \\
        \midrule
        \textbf{Pre-training} & \\
        Batch Size &32 \\
        Optimizer & AdamW \\
        Learning Rate (lr) & 5e-5 \\
        Lr Scheduler & inverse\_sqrt  \\
        Warmup Step & 2,500 \\
        Total Step & 100,000 \\
        Extended Vocabulary Size & 53,843 \\
        Max Sequence Length & 512 \\
        Mix Precision & fp16 (O1) \\
        Number of Parameters &142M (base), 409M (large) \\
        Training Time & 13h (base), 70h (large) \\
        \midrule
        \textbf{Fine-tuning (Parsing)} & \\
        Batch Size & 8 \\
        Optimizer & AdamW \\
        Learning Rate (lr) & 3e-5 (base), 8e-6 (large) \\
        Lr Scheduler & constant  \\
        Warmup Step & 0 \\
        Total Epoch & 20 \\
        Early Stop & 5 \\
        Max Sequence Length & 512 \\
        Beam Size & 5 \\
        Length Penalty & 1.0 \\
        Label Smoothing & 0 \\
        Mix Precision & fp16 (O1) \\
        Training Time & 6h (base), 12h (large) \\
        \midrule
        \textbf{Fine-tuning (Generation)} & \\
        Batch Size & 8 \\
        Optimizer & AdamW \\
        Learning Rate (lr) & 5e-6 (base), 2e-6 (large) \\
        Lr scheduler & constant  \\
        Warmup Step & 0 \\
        Total Epoch & 20 \\
        Early Stop & 5 \\
        Max Sequence Length & 512 \\
        Beam Size & 5 \\
        Length Penalty & 1.0 \\
        Label Smoothing & 0 \\
        Mix Precision & fp16 (O1) \\
        Training Time & 3h (base), 6h (large) \\
        \bottomrule
    \end{tabular}
    \caption{Hyper-parameters of our models on Pre-training and Fine-tuning.}
    \label{tab:params}
\end{table}

Table~\ref{tab:params} lists all model hyper-parameters used for our experiments.
We implement our model based on \textit{Pytorch} and \textit{Huggingface Transformers}.
The pre-processed data, source code and pre-trained models are released at \url{https://github.com/muyeby/AMRBART}.

\section{Fine-grained Evaluation Metric for AMR Parsing}\label{appendix:metrix}
The Smatch score~\cite{cai-knight-2013-smatch} measures the degree of overlap between the gold and the prediction AMR graphs. 
It can be further broken into different sub-metrics, including:
\begin{itemize}
    \item Unlabeled (Unlab.): Smatch score after removing edge-labels
    \item NoWSD: Smatch score after ignoring Propbank senses (\emph{e.g.}, go-01 vs go-02)
    \item Concepts (Con.): $F$-score on the concept identification task
    \item Wikification (Wiki.): $F$-score on the wikification (\texttt{:wiki} roles)
    \item Named Entity Recognition (NER): $F$-score on the named entities (\texttt{:name} roles).
    \item Reentrancy (Reen.): Smatch score on reentrant edges.
    \item Negation (Neg.): $F$-score on the negation detection (\texttt{:polarity} roles).
    \item Semantic Role Labeling (SRL): Smatch score computed on~\texttt{:ARG-i} roles.
\end{itemize}
\section{More Experimental Results}\label{appendix:more-experiments}

\begin{table}
	\centering
	\small
	\begin{tabular}{lcc}
		\toprule
        \textbf{Setting} & AMR parsing & AMR-to-text \\
		\midrule 	
		BART & 82.7 & 42.5 \\
		\quad + silver (fine-tuning) & 82.6 & 44.9 \\
		\quad + silver (denoising)  & 83.6 & 45.6 \\
	\bottomrule
	\end{tabular}
	\caption{Ablation study on silver data and denoising tasks.}
	\label{tab:ablation2}
\end{table}


\subsection{Effect of Unified Framework}\label{appendix:effect}
\begin{figure}
	\centering
	\includegraphics[width=0.4\textwidth]{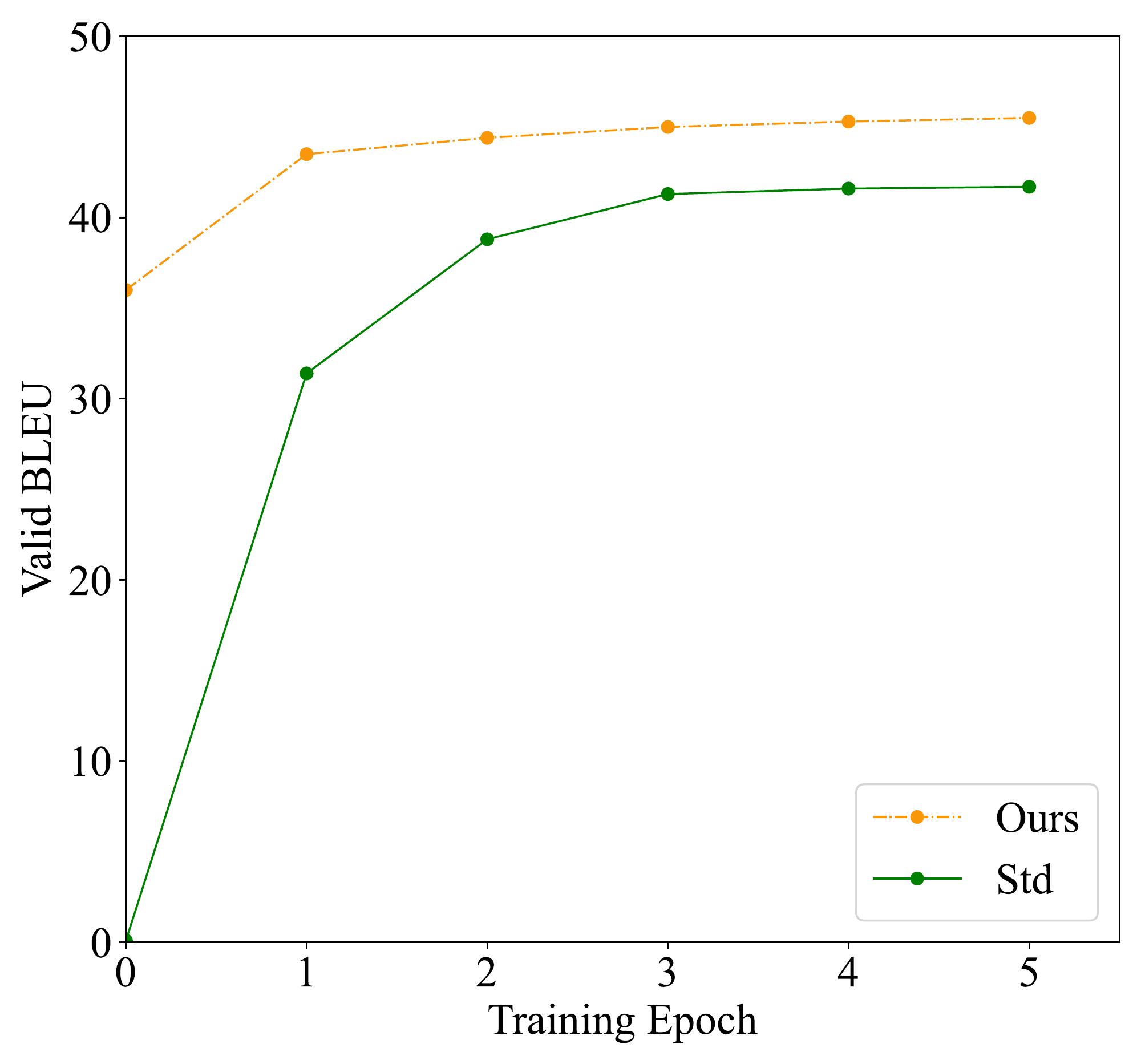}
	\caption{The learning curve of standard framework (std) and our unified framework (ours) on AMR-to-text generation task.}
	\label{fig:curve}
\end{figure}
Figure~\ref{fig:curve} compares the learning curve between our unified framework and standard framework (fine-tuning from vanilla BART, i.e., Bevilacqua+) on AMR2.0 validation set\footnote{We use the same learning rate and optimizer.}.
It can be observed that our system has a initial BLEU score of $36.0$, which is significantly ($p$< 0.001) better than the baseline.
This confirm that our unified framework can reduce the gap between pre-training and fine-tuning.
In addition, the training curve of the proposed model converges faster while the BLEU score is better than the baseline. 
This indicates that our model has a larger capacity than baseline.

\subsection{Impact of denoising Tasks}\label{appendix:denoising}
To distinguish the contribution of de-nosing tasks and silver data, an ablation study is present where we 1) ``fine-tune'' a vanilla BART on silver data following our unified framework (i.e., $\overline{\texttt{t}}\texttt{g}$\texttt{2t} and $\texttt{t}\overline{\texttt{g}}$\texttt{2g}); 2) continue pre-train a BART on silver data according to proposed de-nosing tasks (in Table~\ref{tab:trainingphase}).
As shown in Table~\ref{tab:ablation2}, we observe that using sliver data for fine-tuning leads to a 0.1 Smatch decrease in AMR parsing and 2.4 BLEU increase in AMR-to-text. 
This observation is consistent with previous works~\cite{konstas2017neural,song2018graph,Bevilacqua_Blloshmi_Navigli_2021}.
In addition, using silver data for pre-training gives further improvements on both tasks, with 1.0 Smatch for AMR pasring and 0.7 BLEU for AMR-to-text generation.
This indicates that our de-nosing tasks can help model to better understand silver data.

\begin{table}[!t]
	\centering
	\small
	\begin{tabular}{lcc}
		\toprule
        \textbf{Setting} & AMR parsing & AMR-to-text \\
		\midrule 	
		\textbf{AMR2.0} & &\\
		Ours (large)& 85.4 & 49.8 \\
		\quad + silver  & 85.1 & 49.6 \\
		\midrule 	
		\textbf{AMR3.0} & &\\
		Ours (large)& 84.2 & 49.2 \\
		\quad + silver  & 83.8 & 48.9 \\
	\bottomrule
	\end{tabular}
	\caption{Model performance on AMR2.0 and 3.0 datasets for AMR parsing and AMR-to-text. For AMR parsing, we report Smatch score here, and for AMR-to-text, we report BLEU-4 score here. +silver denotes to that further fine-tuning the model on silver data.}
	\label{tab:silverdata}
\end{table}

\subsection{Are Silver Data Still Helpful for Fine- tuning after Being Used for Pre-training?}\label{appendix:silverdata}

As discussed in Section~\ref{sec:main-res}, we find that graph pre-training is a better way to make use of silver data compared with fine-tuning.
We further investigate whether fine-tuning our model on silver data can still bring improvement.
As shown in Table~\ref{tab:silverdata}, our models achieve the best performance on all tasks and datasets, indicating that further fine-tuning our models on silver data decreases the performance.
This can be that silver data are already presented in the pre-training phase and thus further fine-tuning can bring no improvement.
In addition, fine-tuning can be more sensitive to data quality than pre-training.
When training data contain noise (silver data), fine-tuning on such data can in turn damage the model performance.
\end{document}